\documentclass{article}

\usepackage{arxiv}

\usepackage[utf8]{inputenc}
\usepackage[T1]{fontenc}
\usepackage{microtype}
\usepackage{inconsolata}
\usepackage{graphicx}
\usepackage[table]{xcolor}  % \rowcolor in tables
\usepackage{booktabs}       % \toprule, \midrule, \bottomrule
\usepackage{amsfonts}       % blackboard math symbols
\usepackage{nicefrac}       % compact symbols for 1/2, etc.
\usepackage{amsmath,amssymb}
\usepackage{subcaption}
\usepackage{multirow}       % \multirow
\usepackage{algorithm}
\usepackage{algpseudocode}
\usepackage{enumitem}
\usepackage{placeins}       % \FloatBarrier
\usepackage{float}          % [H] placement for appendix tables
\usepackage{url}
\usepackage[round]{natbib}
\usepackage[colorlinks=true,linkcolor=darkblue,citecolor=darkblue,urlcolor=darkblue]{hyperref}

\newcommand{\arxivversion}{}

\makeatletter
\renewenvironment{table}{\@float{table}}{\end@float}
\makeatother
\renewenvironment{table*}{\begin{table}}{\end{table}}


\definecolor{darkblue}{rgb}{0, 0, 0.5}
\definecolor{deepgreen}{rgb}{0, 0.5, 0}
\renewcommand\cite{\citep}

\title{Self-Distillation for Supervised Fine-Tuning}

\renewcommand{\shorttitle}{Self-Distillation for Supervised Fine-Tuning}

\author{
  Wenxuan Zhu\textsuperscript{1} \quad
  Bang An\textsuperscript{1} \quad
  Shuming Liu\textsuperscript{1} \quad
  Wenqing Cui\textsuperscript{1} \quad
  Tianqi Xu\textsuperscript{1} \\
  \textbf{Bing Li\textsuperscript{1}} \quad
  \textbf{Yibo Yang\textsuperscript{2}} \quad
  \textbf{Peter~Wonka\textsuperscript{1}} \quad
  \textbf{Bernard Ghanem\textsuperscript{1}} \\[1ex]
  \normalfont
  \textsuperscript{1}King Abdullah University of Science and Technology \quad
  \textsuperscript{2}Shanghai Jiao Tong University
}

\date{}

\begin{document}
\maketitle

\begin{abstract}

\noindent Supervised fine-tuning treats the expert demonstration or gold response as a one-hot target, suppressing valid alternative continuations that the model would otherwise assign probability to.
We propose Gold-Conditioned Distillation (GCD), a method that turns expert demonstrations into model-native supervision. 
GCD uses one model as both teacher and student: the teacher conditions on a prompt containing the gold response in context, the student conditions on the original prompt as at inference, and the student is trained to minimize the forward KL divergence to the teacher's next-token distribution at each gold response token.
This lets the model use its prior knowledge to teach itself the new task: correctness is anchored by the gold response, while the soft target preserves plausible alternatives.
Across various training tasks on Qwen and Llama models, GCD improves new-task accuracy over SFT and generally reduces degradation on general benchmarks.
\ifdefined\arxivversion
Code is available on \href{https://github.com/WenxuanZhu1103/gcd}{GitHub}.
\fi

\end{abstract}

\section{Introduction}

Large language models (LLMs) are typically trained once on a massive corpus and then deployed as static artifacts. In practice, however, the world does not hold still. New facts emerge, new tools appear, new domains demand attention---and the model must incorporate this knowledge without forgetting what it already knows. This is the post-training problem, and for modern LLMs it remains a major open challenge.

The standard approach, supervised fine-tuning (SFT), trains the model to imitate the expert demonstrations token by token. This procedure is simple, stable, and widely used. However, it also imposes a rigid supervision interface: every demonstrated token is treated as a one-hot target, and every non-demonstrated token is treated as equally wrong.

For language models, this supervision format can be unnecessarily sharp. A demonstration tells the model which response is correct, but it does not imply that every continuation supported by the model's distribution should be driven to zero. This distinction is important in continual post-training, where a model must acquire new facts, tools, domains, or response formats without unnecessarily destroying the distributional structure that supports its existing capabilities. In this setting, the bottleneck is not only the amount or quality of post-training data, but also the interface through which demonstrations are converted into learning signals.

Recently, \citet{shenfeld2026self} proposed Self-Distillation Fine-Tuning (SDFT), which exploits the in-context learning ability of LLMs for continual post-training. The core idea is that conditioning a model on the gold response in its prompt produces a useful supervision signal. The gold-response-aware model serves as a \emph{teacher} and the gold-response-unaware model as a \emph{student}; SDFT then samples on-policy rollouts from the student in real time and trains the student to match the teacher via KL divergence. 
However, SDFT requires the student to roll out a completion at every training step, which increases training time. Moreover, early in training or on unfamiliar domains these rollouts are often incorrect, and because the teacher and student see different contexts, the teacher's distribution over an erroneous sequence may yield a noisy signal.
More fundamentally, our experiments suggest that SDFT is not effective at learning in a novel domain; like reinforcement learning in LLMs, it is better suited for domains where the model already has some prior capabilities~\cite{yue2504does, wen2025reinforcement}.

Inspired by SDFT's core insight that in-context conditioning on the gold response elicits a usable soft target, we propose \textbf{Gold-Conditioned Distillation (GCD)}, a simple modification that sidesteps the issues above by replacing the student's rollout with the gold response from the dataset. The teacher conditions on a \emph{privileged} prompt that exposes the gold response, while the student conditions only on the question; the student is then trained to match the teacher's token-level distribution evaluated on the \emph{fixed} gold response from the dataset via forward KL divergence loss.

We evaluate GCD under two LLM fine-tuning settings across different models, Llama-3.1-8B-Instruct \cite{grattafiori2024llama} and Qwen2.5-3B-Instruct \cite{Yang2024Qwen25TR} as well as Qwen3-4B-Base and Qwen3-8B-Base \cite{yang2025qwen3}.
The first tests adaptation in partially known domains.
The second training setting tests learning new knowledge from scratch in domains where the model has near-zero initial accuracy.
Across both settings, GCD offers a favorable stability--plasticity trade-off: it often improves new-task acquisition over SFT and usually preserves general capabilities at a comparable or slightly better level.

Our contributions are summarized as:
\begin{itemize}[leftmargin=*]
 \item  We propose Gold-Conditioned Distillation (GCD), a simple yet effective offline self-distillation method for learning from expert demonstrations, achieving strong average new-task learning performance while modestly improving or maintaining retention in most settings. %GCD uses the gold response as privileged context to obtain a model-native soft target, and trains the student to match this target along the gold trajectory.
\item We show that GCD improves held-out generalization in partially known domains and usually preserves general capabilities better than SFT. Token-level analysis shows that the gains come from preserving structured probability mass on model-plausible alternatives rather than from simple label smoothing.
\end{itemize}

\section{Related Work}

\paragraph{Catastrophic forgetting and continual learning.}
Catastrophic forgetting occurs when training on new data reduces a model's performance on previously learned data or tasks \cite{kirkpatrick2017overcoming, mccloskey1989catastrophic, parisi2019continual, van2024continual, DeLange2019ACL, wang2024comprehensivesurveycontinuallearning}. Recent studies show that forgetting is a practical issue in LLMs. Continual instruction tuning can hurt domain knowledge, reasoning, reading comprehension, general ability, and instruction following \cite{luo2025empirical, wang2023trace, zhang2023citb, li2024examiningforgettingcontinualpretraining}. Other work studies the causes of forgetting in LLM tuning and relates it to loss-landscape properties \cite{li2024revisiting, ren2024analyzing}. Continual learning aims to adapt models over time while preserving earlier capabilities. Classical methods are often grouped into replay-based approaches \cite{buzzega2020dark, rebuffi2017icarl, shin2017continual}, regularization-based approaches \cite{kirkpatrick2017overcoming, li2017learning, zenke2017continual}, and architecture-based approaches \cite{serra2018overcoming, mallya2018packnet}. These approaches are harder to apply directly to LLM fine-tuning: replay may require unavailable past data, regularization can add storage and computation costs, and architecture-based methods often require task-specific components. Recent methods also adapt replay and parameter-efficient tuning to LLMs, including synthetic rehearsal \cite{huang2024mitigating, wang2024inscl}, orthogonal LoRA subspaces \cite{qiao2024learn, ren2024analyzing, wang2023orthogonal}, and analyses of why LoRA can forget less \cite{biderman2024lora}. These methods show progress, but they still involve trade-offs between learning new data, preserving old abilities, and adding extra data, computation, or model components.

\paragraph{Self-distillation.}
Knowledge distillation trains a student model to match the behavior of a teacher model \cite{FitNets, Zagoruyko2016PayingMA, Wang2020KnowledgeDA, zhao2022decoupledknowledgedistillation, yang2022maskedgenerativedistillation, chi2023normkdnormalizedlogitsknowledge, Wei2024ScaleDD, song2026surveyonpolicydistillationlarge, cho2019efficacy, hinton2015distilling, Mirzadeh2019ImprovedKD, park2019relational, stanton2021does, tung2019similarity, wang2025feature, zhou2021rethinking}. The teacher signal can be a soft output distribution, intermediate features, attention maps, or contrastive relations \cite{gou2021knowledge, guo2023class, ojha2023knowledge, sun2024logit, tian2019contrastive, Zagoruyko2016PayingMA, wang2025qcrd}. Self-distillation is a special case of knowledge distillation in which the teacher is the same model, a previous version of the model, or the same model under a different input condition. Recent LLM distillation methods study compression \cite{jiao2020tinybert, liu2024ddk}, capability transfer \cite{jiang2025teach, li2025learning, mukherjee2023orca}, and self-improvement \cite{song2025self, yang2024self, zhang2024enhancing}. Since LLMs generate sequences, both the distillation objective and the training distribution matter: MiniLLM studies reverse-KL distillation for LLMs \cite{gu2024minillm}, while Generalized Knowledge Distillation trains on student-generated sequences with teacher feedback to reduce the gap between training and inference \cite{agarwal2024policy}. Closely related to our work, context distillation trains a model to internalize behavior that appears when useful context, examples, or reasoning steps are provided \cite{snell2022learning}. \citet{yang2024self} use model-generated data to reduce the distribution gap in LLM fine-tuning, and SDFT \cite{shenfeld2026self} uses the current model conditioned on a demonstration as its own teacher and generates on-policy completions for training. Our method keeps the same idea of a context-conditioned self-teacher, but uses the gold response as the completion, avoiding online generation and sampling noise.
\section{Method}
\label{sec:method}
\begin{figure*}[t]
    \centering
    \includegraphics[width=0.85\textwidth]{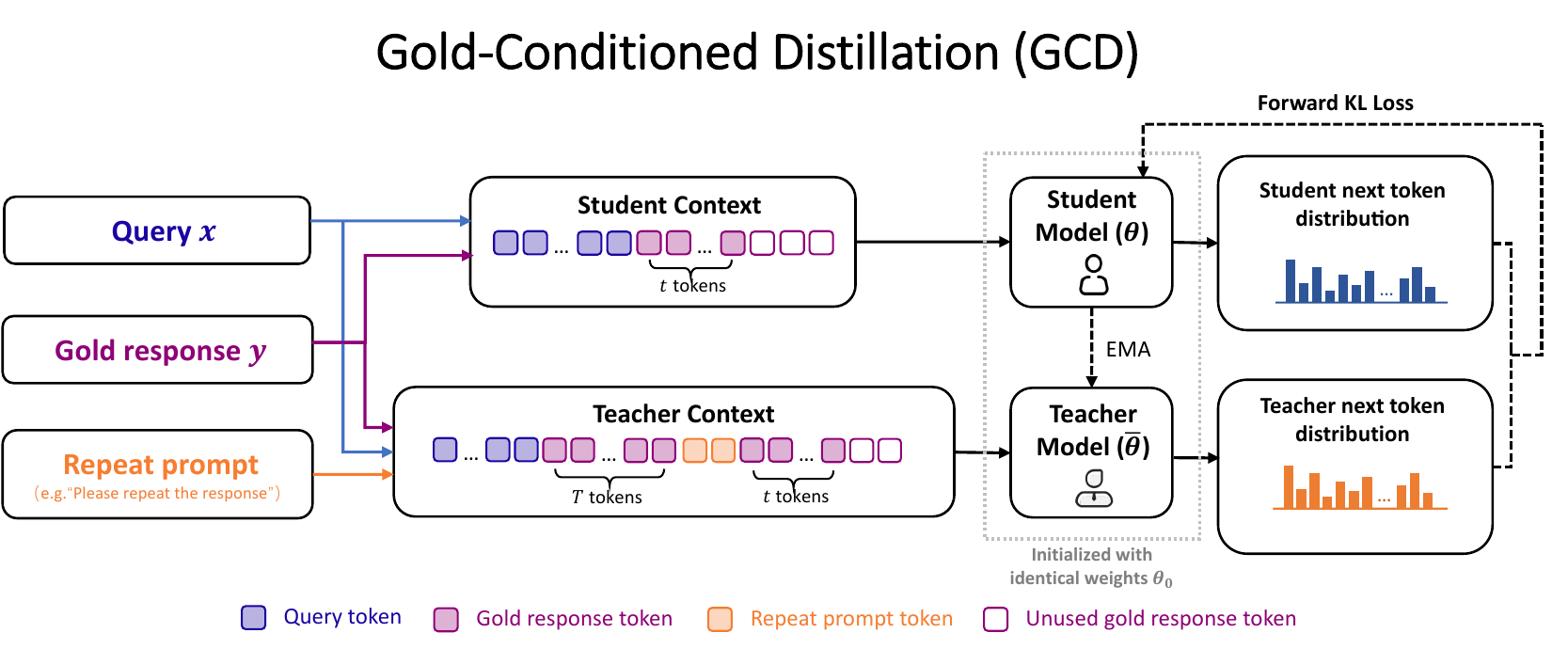}
    \caption{
    Overview of Gold-Conditioned Distillation (GCD).
    The student predicts the next token from the query $x$ and the gold-response prefix $y_{<t}$; the teacher predicts it from a privileged context that additionally contains the full gold response $y$ and a repeat prompt.
    The student is trained with a forward KL loss to match the teacher's gold-conditioned next-token distribution, and the teacher is an exponential moving average of the student.
    }
    \label{fig:gcd_overview}
\end{figure*}

We propose \textbf{Gold-Conditioned Distillation (GCD)}, a token-level
distillation method for learning from expert demonstrations. 
The goal is to replace the one-hot targets used in supervised fine-tuning with model-native targets produced by the model itself under gold response conditioning.

The student and teacher are initialized from the same model, but only the student receives gradient updates; the teacher is maintained as an EMA copy of the student throughout training. Thus, the teacher provides a slowly evolving self-distillation target. GCD is fully offline: neither the student nor the teacher autoregressively generates responses during training; both are evaluated only under teacher forcing along the gold response from the dataset.

\paragraph{Student and teacher contexts.}
Given an input $x$ and its gold response
$y^{*} = (y_{1}^{*}, \dots, y_{T}^{*})$, at position $t$ the student
conditions on the standard teacher-forced prefix, while the teacher conditions
on a \emph{privileged} prefix that additionally exposes the full gold response
through a template $\phi(x, y^{*})$:
\begin{equation}
\begin{aligned}
c_{t}^{(S)} &= [\,x \,;\, y_{<t}^{*}\,],\\
c_{t}^{(T)} &= [\,\phi(x, y^{*}) \,;\, y_{<t}^{*}\,],
\end{aligned}
\end{equation}
where $[\,\cdot\,;\,\cdot\,]$ denotes token-sequence concatenation.
The student and teacher next-token distributions are then obtained by
evaluating each model on its own context:
\begin{equation}
\begin{aligned}
\pi_{S}(\cdot \mid c_{t}^{(S)}) &= \pi_{\theta}(\cdot \mid x,\, y_{<t}^{*}),\\
\pi_{T}(\cdot \mid c_{t}^{(T)}) &= \pi_{\bar{\theta}}(\cdot \mid \phi(x, y^{*}),\, y_{<t}^{*})
\end{aligned}
\end{equation}
where $\theta$ denotes the student parameters, and $\bar{\theta}$ denotes the EMA
teacher parameters. The two models are initialized identically,
$\bar{\theta}_{0} = \theta_{0}$, but diverge during training because only
$\theta$ receives gradient updates.

A concrete instantiation of the privileged template is:
\ifdefined\arxivversion
\begin{equation}
\phi(x, y^{*}) \;=\; x \;\oplus\; \text{``The correct response is:''}
\;\oplus\; y^{*}
\;\oplus\; \text{``Please repeat the correct response.''}
\end{equation}
\else
\begin{equation}
\begin{aligned}
&\phi(x, y^{*}) \;=\; x \;\oplus\; \text{``The correct response is:''} \\
&\quad \oplus\; y^{*} \\
&\quad \oplus\; \text{``Please repeat the correct response.''}
\end{aligned}
\end{equation}
\fi
Both distributions are evaluated on the \emph{same} gold prefix states
$y_{<t}^{*}$; the only asymmetry is that the teacher uses an EMA-smoothed
parameterization and observes an additional gold-revealing segment.

\paragraph{Objective.}
We train the student by matching the privileged teacher distribution with a
token-level forward KL, averaged over positions and over the data distribution
$\mathcal{D}$:
\ifdefined\arxivversion
\begin{equation}
\mathcal{L}(\theta) \;=\;
\mathbb{E}_{(x,\,y^{*}) \sim \mathcal{D}}
\Bigg[\,
\frac{1}{T}\sum_{t=1}^{T}
D_{\mathrm{KL}}\!\Big(
\operatorname{sg}\!\big[\pi_{\bar{\theta}}(\cdot \mid c_{t}^{(T)})\big]
\;\Big\|\;
\pi_{\theta}(\cdot \mid c_{t}^{(S)})
\Big)
\Bigg]
\end{equation}
\else
\begin{equation}
\begin{aligned}
&\mathcal{L}(\theta) \;=\;
\mathbb{E}_{(x,\,y^{*}) \sim \mathcal{D}}
\Bigg[\,
\frac{1}{T}\sum_{t=1}^{T}
D_{\mathrm{KL}}\!\Big( \\
&\qquad
\operatorname{sg}\!\big[\pi_{\bar{\theta}}(\cdot \mid c_{t}^{(T)})\big]
\;\Big\|\;
\pi_{\theta}(\cdot \mid c_{t}^{(S)})
\Big)
\Bigg]
\end{aligned}
\end{equation}
\fi
where $\operatorname{sg}[\cdot]$ denotes the stop-gradient operator. The
teacher distribution is therefore treated as a fixed target for each update,
and gradients are applied only to the student parameters $\theta$. We adopt
the forward, mode-covering KL direction, so that the student is penalized for
placing low probability on tokens that the teacher considers likely under the
gold-revealed context.
An ablation in Appendix~\ref{app:kl_direction_ablation} confirms this choice: replacing it with
reverse KL lowers accuracy on both LiveNews and Tooluse.

\paragraph{EMA teacher update.}
After each student update, the teacher parameters are updated using an
exponential moving average of the student:
\begin{equation}
\bar{\theta} \leftarrow (1-\alpha) \bar{\theta} + \alpha\theta
\end{equation}
where $\alpha \in [0,1]$ controls the update rate. 
In this paper, we choose $\alpha = 0.01$ for the EMA update following SDFT~\cite{shenfeld2026self}.
We conduct an ablation study on $\alpha$ in Appendix~\ref{app:ema_ablation}: GCD is stable for small update
rates ($\alpha \leq 0.05$), while large rates that make the teacher track the student too closely cause training to collapse.

\section{Experiments}
\label{sec:sec_4}
\begin{table}[t]
\centering
\ifdefined\arxivversion
\small
\setlength{\tabcolsep}{10pt}
\else
\fontsize{8pt}{9.5pt}\selectfont   % 8pt so the six columns fit the 7.7cm ACL column
\setlength{\tabcolsep}{4pt}
\fi
\begin{tabular}{l ccccc}
\toprule
\ifdefined\arxivversion
\textbf{Method} & \textbf{Tooluse} & \textbf{Chemistry} & \textbf{Physics \& Material} & \textbf{Biology} & \textbf{Avg.} \\
\else
\textbf{Method} & \textbf{Tooluse} & \textbf{Chemistry} & \shortstack[c]{\textbf{Physics \&}\\\textbf{Material}} & \textbf{Biology} & \textbf{Avg.} \\
\fi
\midrule
\rowcolor{gray!12}
\multicolumn{6}{l}{\textbf{\textsc{Qwen2.5-3B-Instruct}}} \\
\midrule
\textcolor{gray}{Base} & \textcolor{gray}{34.02} & \textcolor{gray}{30.77} & \textcolor{gray}{44.83} & \textcolor{gray}{76.67} & \textcolor{gray}{46.57} \\
SFT           & \underline{60.83} & \underline{65.09} & \textbf{63.79}    & \underline{85.94} & \underline{68.91} \\
SDFT          & 55.67             & 63.31             & 52.30             & 83.53             & 66.38 \\
\textbf{GCD} & \textbf{66.30}    & \textbf{66.67}    & \underline{63.22} & \textbf{86.45}    & \textbf{70.66} \\
\midrule
\rowcolor{gray!12}
\multicolumn{6}{l}{\textbf{\textsc{Qwen3-4B-Base}}} \\
\midrule
\textcolor{gray}{Base} & \textcolor{gray}{36.08} & \textcolor{gray}{38.46} & \textcolor{gray}{56.32} & \textcolor{gray}{84.56} & \textcolor{gray}{53.86} \\
SFT           & \underline{62.89} & \textbf{70.02}    & \textbf{68.40}    & \underline{86.11} & \underline{71.86} \\
SDFT          & 60.83             & 54.24             & 60.92             & 78.04             & 63.51 \\
\textbf{GCD} & \textbf{71.13}    & \underline{69.43} & \underline{67.24} & \textbf{86.45}    & \textbf{73.56} \\
\midrule
\rowcolor{gray!12}
\multicolumn{6}{l}{\textbf{\textsc{Llama-3.1-8B-Instruct}}} \\
\midrule
\textcolor{gray}{Base} & \textcolor{gray}{52.57} & \textcolor{gray}{30.18} & \textcolor{gray}{38.51} & \textcolor{gray}{70.84} & \textcolor{gray}{48.03} \\
SFT           & \underline{65.98} & \underline{68.64} & \underline{60.35} & \textbf{86.97}    & \underline{70.49} \\
SDFT          & 35.05             & 39.65             & 47.13             & 43.91             & 41.44 \\
\textbf{GCD} & \textbf{67.01}    & \textbf{68.84}    & \textbf{62.64}    & \underline{85.94} & \textbf{71.11} \\
\midrule
\rowcolor{gray!12}
\multicolumn{6}{l}{\textbf{\textsc{Qwen3-8B-Base}}} \\
\midrule
\textcolor{gray}{Base} & \textcolor{gray}{55.67} & \textcolor{gray}{39.25} & \textcolor{gray}{55.75} & \textcolor{gray}{83.36} & \textcolor{gray}{58.51} \\
SFT           & 62.89             & \textbf{69.82}    & 64.94             & \textbf{86.96}    & \underline{71.15} \\
SDFT          & \underline{63.92} & 59.57             & \underline{66.67} & 85.76             & 68.98 \\
\textbf{GCD} & \textbf{64.95}    & \underline{69.63} & \textbf{68.97}    & \underline{86.79} & \textbf{72.59} \\
\bottomrule
\end{tabular}
\caption{\label{tab:exp1_learn}
  New-task accuracy (\%) across four downstream tasks and four backbone models.
  \textbf{Bold} and \underline{underline} mark the best and second-best
  among SFT / SDFT / GCD per (model, task) cell (Base excluded).}
\end{table}
\subsection{Experiment Settings}
\textbf{Models.}
We evaluate four pretrained language models spanning model families, scales, and
training regimes: Qwen2.5-3B-Instruct, Qwen3-4B-Base, Llama-3.1-8B-Instruct, and Qwen3-8B-Base.
The base models are included to test whether GCD's gold-conditioned teacher signal remains effective in the absence of post-training instruction-following priors.

\textbf{Dataset and Evaluation.}
We conduct fine-tuning experiments under two learning settings, distinguished by the model's initial capability in the target domain.
\begin{itemize}[leftmargin=*]
    \item \textbf{Partial-knowledge setting.}
    In this setting, the model already possesses partial ability in the target domain before fine-tuning. We fine-tune and evaluate on four datasets: Tooluse~\cite{Tang2023ToolAlpacaGT}, which requires generating the correct API call and JSON arguments in response to user requests; and three scientific multiple-choice datasets, namely Chemistry, Physics \& Material (physics and materials science), and Biology.
    \item \textbf{Near-zero-knowledge setting.}
    In this setting, the model has almost no prior knowledge of the target domain, with initial accuracy close to zero on LiveNews~\cite{zhang2026livenewsbench}. We fine-tune the model on LiveNews to evaluate its ability to continuously acquire new knowledge through updates. LiveNews consists of news question-answering instances collected from September 2025 and January 2026, which post-date the models' training cutoff.
\end{itemize}
We evaluate models from two perspectives:
(a) \textbf{new-task performance}, measured by accuracy on held-out evaluation splits of the corresponding fine-tuning datasets; and
(b) \textbf{previous capability retention}, measured on five general benchmarks (MMLU, HellaSwag, Winogrande, TruthfulQA, and IFEval) using \texttt{lm-evaluation-harness} (settings in Appendix~\ref{app:lmeval}). These benchmarks collectively assess knowledge, commonsense reasoning, language understanding, factuality, and instruction-following ability.
For the new tasks, all responses are generated with greedy decoding; LiveNews answers are graded by Qwen3.5-9B, also with greedy decoding.
Please refer to Appendix~\ref{app:datasets} for more details on the dataset setup.

\textbf{Baselines.}
SFT: standard supervised fine-tuning.
SDFT~\cite{shenfeld2026self}: the original self-distillation fine-tuning
method, in which the student generates an on-policy response and the
gold-conditioned teacher provides the soft target along that
generated response (see Sec.~\ref{sec:method}). Following a known issue in the released implementation, we compute its KL in the forward direction (Appendix~\ref{app:exp_details}).

\textbf{Hyperparameters.}
For Qwen2.5-3B-Instruct and Qwen3-4B-Base, we conduct full fine-tuning using a learning rate of \(5\times10^{-5}\). For Llama-3.1-8B-Instruct and Qwen3-8B-Base, we adopt LoRA~\cite{hu2021lora} with learning rates of \(1\times10^{-4}\) and \(4\times10^{-4}\), respectively. Across all three training methods (SFT, SDFT, and GCD), we keep the other variables like optimizer, learning-rate scheduler, batch size, sequence length, and number of training epochs the same for each \((\text{model}, \text{task})\) setting. Unless otherwise specified, all evaluation results are reported using the final checkpoint obtained at the end of training.

% ---------------------------------------------------------------
\subsection{Results in the Partial-Knowledge Setting}
\label{sec:exp_1}

\begin{table*}[t]
\centering
% \small
\setlength{\tabcolsep}{5pt}
\resizebox{\textwidth}{!}{%
\begin{tabular}{ll|c|cccccc}
\toprule
\multirow{2}{*}{\textbf{Model}} & \multirow{2}{*}{\textbf{Method}} & \multirow{2}{*}{\textbf{New Task}} & \multicolumn{5}{c}{\textbf{Previous Tasks}} & \multirow{2}{*}{\textbf{Avg.}} \\
\cmidrule(lr){4-8}
& & & MMLU & HellaSwag & Winogrande & TruthfulQA & IFEval & \\
\midrule
\rowcolor{gray!12}
\multicolumn{9}{l}{\textbf{\textsc{New Task: Tooluse}}} \\
\midrule
\multirow{4}{*}{Qwen2.5-3B-Instruct}
 & \textcolor{gray}{Base} & \textcolor{gray}{34.02} & \textcolor{gray}{65.45} & \textcolor{gray}{74.85} & \textcolor{gray}{69.69} & \textcolor{gray}{58.63} & \textcolor{gray}{53.33} & \textcolor{gray}{64.39} \\
& SFT           & \underline{60.83} & 64.39             & 72.41             & 69.14             & 52.17             & \underline{40.28} & 59.68 \\
& SDFT          & 55.67             & \textbf{65.10}    & \textbf{74.22}    & \underline{69.22} & \textbf{58.65}    & \textbf{41.22}    & \textbf{61.68} \\
& \textbf{GCD} & \textbf{66.30}    & \underline{64.71} & \underline{72.62} & \textbf{69.93}    & \underline{52.84} & 38.49             & \underline{59.72} \\
\cmidrule(lr){1-9}
\multirow{4}{*}{Qwen3-4B-Base}
 & \textcolor{gray}{Base} & \textcolor{gray}{36.08} & \textcolor{gray}{71.22} & \textcolor{gray}{73.68} & \textcolor{gray}{70.64} & \textcolor{gray}{53.46} & \textcolor{gray}{41.08} & \textcolor{gray}{62.02} \\
 & SFT           & \underline{62.89} & 68.79             & 69.54             & 68.43             & 49.22             & \underline{29.16} & 57.03 \\
 & SDFT          & 60.83             & \textbf{70.62}    & \textbf{72.84}    & \textbf{71.51}    & \textbf{53.64}    & \textbf{33.41}    & \textbf{60.40} \\
 & \textbf{GCD} & \textbf{71.13}    & \underline{70.05} & \underline{71.91} & \underline{69.85} & \underline{50.95} & 25.71             & \underline{57.69} \\
\cmidrule(lr){1-9}
\multirow{4}{*}{Llama-3.1-8B-Instruct}
 & \textcolor{gray}{Base} & \textcolor{gray}{52.57} & \textcolor{gray}{68.27} & \textcolor{gray}{79.52} & \textcolor{gray}{73.64} & \textcolor{gray}{54.52} & \textcolor{gray}{53.96} & \textcolor{gray}{65.98} \\
 & SFT           & \underline{65.98} & 67.74             & 77.97             & \textbf{73.95}    & 54.09             & 46.11             & 63.97 \\
 & SDFT          & 35.05             & \textbf{68.19}    & \textbf{79.47}    & \textbf{73.95}    & \underline{54.71} & \textbf{53.59}    & \textbf{65.98} \\
 & \textbf{GCD} & \textbf{67.01}    & \underline{68.03} & \underline{79.11} & \textbf{73.95}    & \textbf{55.02}    & \underline{47.79} & \underline{64.78} \\
\cmidrule(lr){1-9}
\multirow{4}{*}{Qwen3-8B-Base}
 & \textcolor{gray}{Base} & \textcolor{gray}{55.67} & \textcolor{gray}{74.55} & \textcolor{gray}{78.59} & \textcolor{gray}{71.74} & \textcolor{gray}{52.28} & \textcolor{gray}{40.76} & \textcolor{gray}{63.58} \\
 & SFT           & 62.89             & 72.98             & 76.37             & 70.48             & \underline{50.82} & \textbf{33.08}    & \underline{60.75} \\
 & SDFT          & \underline{63.92} & \textbf{73.86}    & \textbf{77.70}    & \textbf{73.09}    & \textbf{53.67}    & 17.12             & 59.09 \\
 & \textbf{GCD} & \textbf{64.95}    & \underline{73.22} & \underline{77.67} & \underline{72.45} & 50.28             & \underline{32.16} & \textbf{61.16} \\
\bottomrule
\end{tabular}%
}
\caption{\label{tab:exp1_forget}
  Previous capability retention on five
  general benchmarks after fine-tuning on Tooluse (\%). \textbf{Bold} and
  \underline{underline} denote the best and second-best results among
  SFT / SDFT / GCD per (model, column) cell (Base excluded).}
\end{table*}

Table~\ref{tab:exp1_learn} reports new-task accuracy on each of the four tasks for all four models. Overall, our method GCD consistently achieves the best average performance across all four models, demonstrating its strong generalization ability on new tasks. Table~\ref{tab:exp1_forget} shows that these new-task gains do not require a larger loss of general capabilities. On Tooluse fine-tuning, GCD improves new-task accuracy over SFT for all four models, while maintaining comparable average performance on previous-task benchmarks. 
We attribute these gains to the model-native soft labels produced by the gold-conditioned teacher. Compared with the one-hot targets used by SFT, these soft labels provide a less brittle supervision signal: the gold token still anchors the correct response, while the remaining probability mass preserves model-plausible alternatives. This encourages the student to learn a smoother local distribution around the expert demonstration, which can improve generalization to held-out examples.

\subsection{Results in the Near-Zero-Knowledge Setting}
\begin{figure*}[t]
    \centering
    \includegraphics[width=\textwidth]{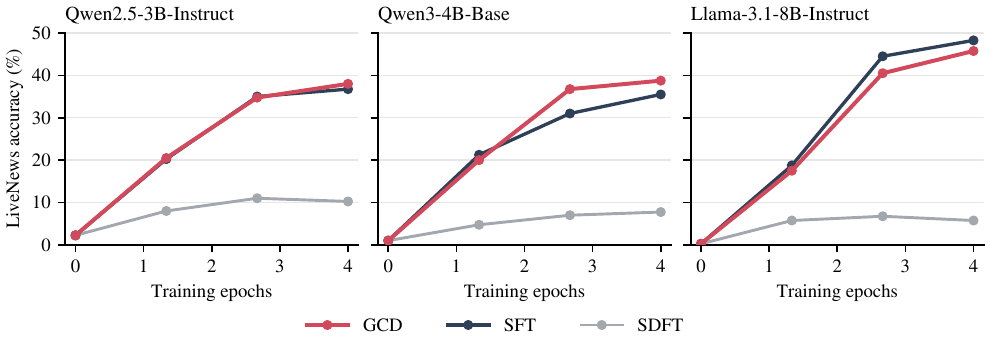}
    \caption{
        \textbf{Learning New Knowledge from Scratch.}
        We report evaluation accuracy as a function of training epochs for
         Qwen2.5-3B-Instruct, Qwen3-4B-Base and Llama-3.1-8B-Instruct 
        on LiveNews dataset.
    }
    \label{fig:exp2}
\end{figure*}
\begin{table*}[t]
\centering
\small
\setlength{\tabcolsep}{5pt}
\resizebox{\textwidth}{!}{%
\begin{tabular}{ll|c|cccccc}
\toprule
\multirow{2}{*}{\textbf{Model}} & \multirow{2}{*}{\textbf{Method}} & \multirow{2}{*}{\textbf{New Task}} & \multicolumn{5}{c}{\textbf{Previous Tasks}} & \multirow{2}{*}{\textbf{Avg.}} \\
\cmidrule(lr){4-8}
& & & MMLU & HellaSwag & Winogrande & TruthfulQA & IFEval & \\
\midrule
\rowcolor{gray!12}
\multicolumn{9}{l}{\textbf{\textsc{New Task: LiveNews}}} \\
\midrule
\multirow{4}{*}{Qwen2.5-3B-Instruct}
 & \textcolor{gray}{Base} & \textcolor{gray}{1.50} & \textcolor{gray}{65.45} & \textcolor{gray}{74.85} & \textcolor{gray}{69.69} & \textcolor{gray}{58.63} & \textcolor{gray}{53.33} & \textcolor{gray}{64.39} \\
 & SFT           & \underline{36.75}    & 64.86             & \underline{74.37} & 67.80             & \underline{36.78} & \underline{43.41} & 57.44 \\
 & SDFT          & 10.25             & \underline{65.12} & \textbf{74.80}    & \textbf{69.69}    & \textbf{53.94}    & 35.12             & \textbf{59.73} \\
 & \textbf{GCD} & \textbf{38.00} & \textbf{65.19}    & 74.24             & \underline{69.06} & 36.54             & \textbf{44.48}    & \underline{57.90} \\
\cmidrule(lr){1-9}
\multirow{4}{*}{Qwen3-4B-Base}
 & \textcolor{gray}{Base} & \textcolor{gray}{1.00} & \textcolor{gray}{71.22} & \textcolor{gray}{73.68} & \textcolor{gray}{70.64} & \textcolor{gray}{53.46} & \textcolor{gray}{41.08} & \textcolor{gray}{62.02} \\
 & SFT           & \underline{35.50} & \underline{67.13} & 72.40             & \underline{70.09} & 36.46             & \textbf{27.31}    & \underline{54.68} \\
 & SDFT          & 7.75              & \textbf{68.07}    & \textbf{73.14}    & 69.53             & \textbf{54.00}    & 23.02             & \textbf{57.55} \\
 & \textbf{GCD} & \textbf{38.75}    & 66.83             & \underline{72.45} & \textbf{70.72}    & \underline{36.83} & \underline{25.00} & 54.36 \\
\cmidrule(lr){1-9}
\multirow{4}{*}{Llama-3.1-8B-Instruct}
 & \textcolor{gray}{Base} & \textcolor{gray}{0.25} & \textcolor{gray}{68.27} & \textcolor{gray}{79.52} & \textcolor{gray}{73.64} & \textcolor{gray}{54.52} & \textcolor{gray}{53.96} & \textcolor{gray}{65.98} \\
 & SFT           & \textbf{48.25}    & 66.27             & 77.99             & 71.90             & 38.48             & 47.78             & 60.48 \\
 & SDFT          & 5.75              & \textbf{67.81}    & \textbf{79.63}    & \textbf{74.74}    & \textbf{53.44}    & \underline{50.10} & \textbf{65.14} \\
 & \textbf{GCD} & \underline{45.75} & \underline{67.02} & \underline{78.96} & \underline{72.61} & \underline{40.21} & \textbf{50.61}    & \underline{61.88} \\
\bottomrule
\end{tabular}%
}
\caption{\label{tab:exp2_forget}
  Previous capability retention on five general-purpose benchmarks after
  fine-tuning on LiveNews (\%). The gray Base row gives pre-fine-tuning
  performance. \textbf{Bold} and \underline{underline} denote the best and
  second-best among SFT / SDFT / GCD per (model, column) cell (Base excluded).}
\end{table*}

\paragraph{Learning new knowledge from scratch.}
Figure~\ref{fig:exp2} reports evaluation accuracy across training epochs on LiveNews. 
In this setting, the base models have near-zero initial accuracy, indicating that the target knowledge is largely absent before fine-tuning. 
We observe that GCD learns the new task as effectively as, and sometimes better than, standard SFT. 
On LiveNews, GCD achieves higher final new-task accuracy than SFT on Qwen2.5-3B-Instruct and Qwen3-4B-Base, while SFT is slightly better on Llama-3.1-8B-Instruct. 
By contrast, SDFT performs poorly on new-task accuracy across all three models. 
This is expected in the from-scratch knowledge-injection setting: since SDFT relies on student-generated rollouts, the early trajectories are often incorrect when the model has little prior knowledge about the target domain. 
The teacher is then asked to provide supervision on low-quality trajectories, leading to weak acquisition of the new knowledge.

\paragraph{Capability retention in knowledge injection.}
Table~\ref{tab:exp2_forget} reports retention on five general-purpose benchmarks after fine-tuning on LiveNews, and shows a clear trade-off between acquiring the new task and preserving previous capabilities.
SDFT obtains the highest previous-task average on all three models, but only because it barely learns the injected knowledge (new-task accuracy below $11\%$), so high retention alone is not useful.
GCD is more balanced: it matches or exceeds SFT on the new task on two of the three models and is slightly below it on Llama-3.1-8B-Instruct, while its previous-task average is higher than SFT's on two models and marginally lower on Qwen3-4B-Base.
The retention gains are modest, and GCD does not fully solve catastrophic forgetting; the takeaway is that GCD reaches SFT-level new-task learning while usually preserving general capabilities slightly better.

\section{Analysis}
\subsection{What the GCD Soft Label Encodes and What the Student Inherits}
\label{sec:interpretability}

GCD conditions the teacher on the gold response, asks it to
reproduce that response token by token, and trains the student under
$D_\mathrm{KL}(\pi_T\,\|\,\pi_\theta)$. We ask two questions empirically:
\textbf{(i)} what does the gold-conditioned target distribution
$\pi_T$ encode beyond the SFT one-hot label, and
\textbf{(ii)} does the trained student inherit that structure at
inference time, when the gold response is no longer in context.

\paragraph{Setup.}
We use Qwen2.5-3B-Instruct on Tooluse dataset. We forward about $200$ training
prompts through the initial model. At every token position $t$ we record the teacher distribution
$\pi_T(\cdot \mid c_t^{(T)})$ under the gold-conditioned context and, as a
control, the distribution of the same model under the student context
$c_t^{(S)}$, i.e., without the gold response; we refer to the latter as the
\emph{base model}. We focus on the \emph{low-confidence gold positions}, selected by the
threshold $\pi_T(y_t^{*}) \le 0.85$ after discarding formatting, whitespace,
and BPE-segmentation artifacts.
% All statistics below are averaged over this set, and the same set is used to evaluate the trained students.

\paragraph{Context-conditional alternatives.}
Representative low-confidence positions place mass on paraphrase
choices, semantically related action verbs, and reasoning-step discourse markers ---
\texttt{`weight'} / \texttt{`mass'},
\texttt{`create'} / \texttt{`get'},
\texttt{`impacted'} / \texttt{`influenced'},
\texttt{`Next'} / \texttt{`Thought'}. Crucially, the choice depends on the
context. To quantify this, we define the \emph{assignment entropy} of a gold
token as the entropy of the distribution of the teacher's top-$1$ alternative
across the low-confidence instances of that token. Across $41$ instances of
\texttt{`Next'}, the top-$1$ alternative takes seven distinct values with an
assignment entropy of $1.41$ nats. Any context-independent regularizer
(uniform or per-token-id label smoothing) has assignment entropy zero by
construction.

\paragraph{Aligned with the base model's natural continuations.}
The Teacher column of Table~\ref{tab:token_stats} compares the teacher's
top-$5$ alternatives with the base model's top-$5$ at the same position: the
average Jaccard overlap is $0.26$, against a random baseline of about
$1.9 \times 10^{-5}$, and $46\%$ of $\pi_T$'s residual non-gold mass lies on
tokens that are also among the base model's top-$5$.
The gold-conditioned residual is therefore not a prompt-template artifact: it
is concentrated on the branch points the base model would explore on its own,
re-weighted toward the gold response.

\begin{table}[t]
\centering
\small
\setlength{\tabcolsep}{4pt}
\begin{tabular}{lcccc}
\toprule
& Base & SFT & GCD & Teacher $\pi_T$ \\
\midrule
$\pi(\text{gold})$          & 0.10 & \textbf{0.94} & 0.86          & 0.55 \\
Mass on teacher top-5       & 0.62 & 0.05          & \textbf{0.13} & 0.45 \\
Mass on base top-5          & 0.88 & 0.05          & \textbf{0.10} & 0.46 \\
Top-5 Jaccard w/ base       & 0.99 & 0.14          & \textbf{0.27} & 0.26 \\
\bottomrule
\end{tabular}
\caption{\label{tab:token_stats}
Token-level distribution statistics at low-confidence gold positions on Tooluse (Qwen2.5-3B-Instruct).
Compared with SFT, GCD assigns less saturated probability to the gold token while preserving substantially more mass on both the gold-conditioned teacher's top alternatives and the base model's natural continuations.
}
\end{table}

% \begin{figure}[t]
%     \centering
%     \includegraphics[width=0.82\linewidth]{figs/src/ls_tooluse_v2.pdf}
%     \caption{
%     Comparison between GCD and label smoothing on Tooluse dataset using Qwen2.5-3B-Instruct.
%     We compare SFT without label smoothing, SFT with different smoothing coefficients, and GCD under the same training setup.
%     GCD consistently achieves higher Tooluse accuracy than all label-smoothing variants, showing that its gain is not simply due to using softer labels.
%     }
%     \label{fig:ls_tooluse}
% \end{figure}

\begin{figure*}[t]
    \centering

    \begin{subfigure}[t]{0.44\linewidth}
        \centering
        \includegraphics[width=\linewidth]{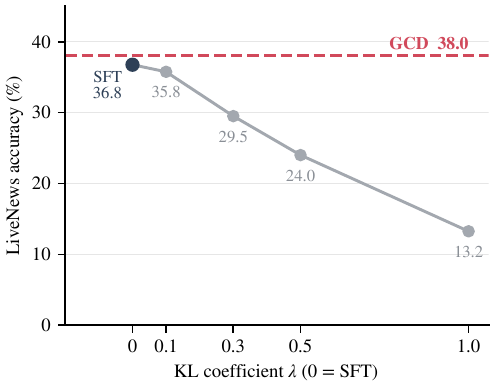}
        \caption{
        Comparison between GCD and SFT with KL regularization toward the base model on LiveNews using Qwen2.5-3B-Instruct.
        We sweep $\lambda \in \{0.1,0.3,0.5,1.0\}$ for SFT+KL(base).
        }
        \label{fig:base-kl}
    \end{subfigure}
    \hfill
    \begin{subfigure}[t]{0.44\linewidth}
        \centering
        \includegraphics[width=\linewidth]{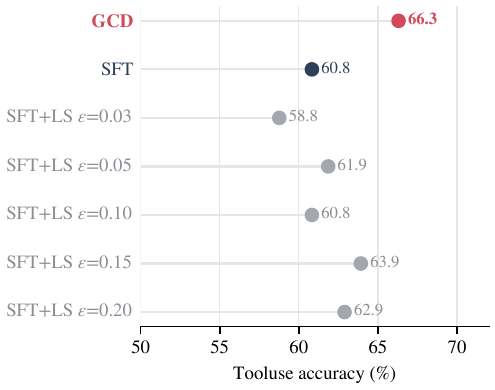}
        \caption{
        Comparison between GCD and label smoothing on Tooluse dataset using Qwen2.5-3B-Instruct.
        We compare SFT without label smoothing, SFT with different smoothing coefficients, and GCD under the same training setup.
        }
        \label{fig:ls_tooluse}
    \end{subfigure}

    \caption{Comparison of GCD with alternative regularization or soft-label baselines.}
    \label{fig:gcd-ablation-side-by-side}
\end{figure*}

\paragraph{Anti-saturation, alternative preservation, base-aligned shape.}
Table~\ref{tab:token_stats} also reports the output distributions of the SFT
and GCD checkpoints at the same low-confidence positions under the student
context. SFT pushes $\pi_\theta(\text{gold})$ to $0.94$, whereas GCD stops at $0.86$.
This is consistent with the direction predicted by the forward-KL objective,
whose fixed point $\pi_\theta = \pi_T$ places only $0.55$ on gold at these
positions; the student does not reach this value because it is trained
against the EMA teacher, which moves toward the student during training,
whereas Table~\ref{tab:token_stats} reports the target of the initial teacher.
GCD retains $2.6\times$ and $2.0\times$ the mass of SFT on the teacher's and
the base model's top alternatives, respectively, and at $10\%$ of the
positions it even ranks an alternative above gold, mirroring positions where
$\pi_T$ itself does. The top-$5$ Jaccard overlap with the base model is
$1.9\times$ that of SFT ($0.27$ vs.\ $0.14$).

\paragraph{Connection to forgetting.}
Because $\pi_T$'s residual mass falls on tokens the base model would
naturally produce, the forward-KL fixed point $\pi_\theta = \pi_T$ pins the
student close to the base distribution at these branch points, whereas the
SFT one-hot loss keeps pushing alternative probability toward zero across
epochs. We hypothesize that this base-aligned shape is a local signature of
the retention differences measured in Sec.~\ref{sec:sec_4}; the gains there
are modest, and establishing the link directly is left for future work.

\subsection{Comparison with SFT+KL}

To test whether the benefit of GCD can be reproduced by simply regularizing toward the base model, we compare it with SFT+KL(base), which augments the SFT objective with a KL penalty toward the base model distribution:
\ifdefined\arxivversion
\begin{equation}
\mathcal{L}
=
\mathcal{L}_{\mathrm{SFT}}
+
\frac{\lambda}{T}\sum_{t=1}^{T}
D_{\mathrm{KL}}\Big(
\pi_{\mathrm{base}}(\cdot \mid x, y^{*}_{<t})
\;\Big\|\;
\pi_{\theta}(\cdot \mid x, y^{*}_{<t})
\Big)
\end{equation}
\else
\begin{equation}
\begin{aligned}
&\mathcal{L}
=
\mathcal{L}_{\mathrm{SFT}}
+
\frac{\lambda}{T}\sum_{t=1}^{T}
D_{\mathrm{KL}}\Big( \\
&\qquad
\pi_{\mathrm{base}}(\cdot \mid x, y^{*}_{<t})
\;\Big\|\;
\pi_{\theta}(\cdot \mid x, y^{*}_{<t})
\Big)
\end{aligned}
\end{equation}
\fi
We sweep the regularization strength $\lambda \in \{0.1, 0.3, 0.5, 1.0\}$ and compare all variants with SFT and GCD on LiveNews with Qwen2.5-3B-Instruct.

As shown in Figure~\ref{fig:base-kl}, KL regularization toward the base model does not improve new-knowledge acquisition: $\lambda=0.1$ behaves like SFT and still trails GCD at the end of training, while $\lambda \ge 0.3$ increasingly constrains the model to the base distribution and falls substantially below SFT. GCD's new-task gains therefore cannot be explained by simply keeping the fine-tuned model close to the base model; note that this comparison concerns new-task accuracy only.

\subsection{Comparison with SFT+Label Smoothing}
\label{sec:label_smoothing}

To test whether the structure of the GCD target matters, or whether any softened version of the one-hot target suffices, we compare against uniformly label-smoothed SFT (SFT+LS), which replaces the one-hot target at each position with
\begin{equation}
q_{\mathrm{LS}}(v)
=
(1-\epsilon)\mathbf{1}[v=y_t^\ast]
+
\frac{\epsilon}{|\mathcal{V}|}
\end{equation}
where $y_t^\ast$ is the gold token, $\epsilon$ is the smoothing coefficient, and $|\mathcal{V}|$ is the vocabulary size. This baseline controls for the effect of softening the target distribution while removing the model- and context-dependent structure present in GCD. We sweep $\epsilon \in \{0.03, 0.05, 0.10, 0.15, 0.20\}$ on Tooluse with Qwen2.5-3B-Instruct.

% Empirically, SFT+LS performs worse than vanilla SFT in our continual learning experiments. This result suggests that naively reducing confidence on the gold token is not sufficient for mitigating forgetting, and can even weaken new-task acquisition.

As shown in Figure~\ref{fig:ls_tooluse}, label smoothing improves over SFT inconsistently and by at most about three points, with some coefficients falling below SFT, whereas GCD achieves the best held-out Tooluse accuracy. The gain of GCD therefore cannot be explained only by increasing the entropy of the target distribution.

\section{Conclusion}

In this paper, we propose GCD, a novel distillation framework for LLM fine-tuning. 
Unlike SFT, which relies on one-hot targets, a gold-conditioned teacher is introduced to provide soft, model-native supervision, enabling the student model to preserve plausible alternatives aligned with the model's prior distribution while improving adaptation to new tasks. Unlike on-policy self-distillation methods such as SDFT, GCD performs distillation directly along the gold response, avoiding costly online rollout generation.  
Extensive experiments show that GCD achieves strong average new-task performance across multiple models and tasks. Compared with SFT, GCD often improves new-task accuracy and provides slightly better retention. Compared with SDFT, GCD substantially improves new-task acquisition in the settings we study.

\section*{Limitations}
This work is an initial exploration of gold-conditioned supervision. In our experiments, we use a simple prompt that asks the teacher to repeat the provided gold response. This design may not fully exploit the model's ability to interpret the demonstration. A better prompt could ask the teacher to compare the gold response with plausible alternatives, identify key decision points, or produce a more calibrated distribution over possible continuations. Such prompts may elicit richer and more informative soft targets than simple repetition.
In addition, our token-level analysis is conducted on a single model and dataset, the retention gains over SFT are modest, the near-zero-knowledge setting covers three of the four models, and all models have at most 8B parameters; whether these findings transfer to larger models remains open.

\section*{Acknowledgement}
The research reported in this publication was supported by funding from King Abdullah University of Science and Technology (KAUST) - Center of Excellence for Generative AI, under award number 5940.

% Keep floats out of the reference list.
\FloatBarrier
\bibliographystyle{plainnat}
\bibliography{references}

\clearpage
\appendix
\section{Dataset Details}
\label{app:datasets}

We evaluate our method on five datasets.
Table~\ref{tab:dataset_stats} provides an overview of the dataset; per-dataset descriptions follow.

\ifdefined\arxivversion\begin{table*}[H]\else\begin{table*}[t]\fi
\centering
\small
\begin{tabular}{lllrr}
\toprule
\textbf{Dataset} & \textbf{Domain} & \textbf{Answer format} & \textbf{\#Train} & \textbf{\#Eval} \\
\midrule
Tooluse              & API tool calling                      & API name + JSON args & 4{,}046 & 97 \\
Chemistry            & Mixed scientific QA (chemistry-heavy) & 4-option MCQ & 2{,}674 & 507 \\
Physics \& Material   & Physics + materials science           & 4-option MCQ & 1{,}257 & 174 \\
Biology              & School- and graduate-level biology    & 2/3/4-option MCQ & 1{,}107 & 583 \\
LiveNews             & Post-training-cutoff news QA          & Free-form answer & 400 & 400 \\
\bottomrule
\end{tabular}
\caption{Dataset overview. \#Train and \#Eval give the number of training and held-out evaluation examples.}
\label{tab:dataset_stats}
\end{table*}

\paragraph{Tooluse.}
Same dataset used in \citet{shenfeld2026self}.
A function-calling benchmark in which the model receives a natural-language
instruction together with the documentation of one or more API tools and must
emit a structured \texttt{Action:<tool\_name>} and \texttt{Action Input:<JSON>} sequence. Each instance contains the tool's name, description,
natural-language documentation, a user instruction, and a gold response
consisting of one or more \texttt{(Action, Action\_Input)} pairs. 

\paragraph{Chemistry.}
Same dataset used in \citet{shenfeld2026self}.
A Chemistry QA corpus in which each example pairs a multiple-choice
question with a chain-of-thought solution
wrapped in \texttt{<reasoning>...</reasoning>}\allowbreak\texttt{<answer>...</answer>} tags. 

\paragraph{Physics \& Material.}
A merged physics + materials-science split derived from
SciKnowEval~\cite{feng2024sciknoweval}.  
Each prompt presents a quantitative question (e.g., Gauss-law fluxes, gravitational lensing, materials
properties) together with the relevant physical constants, and each gold response is a worked solution followed by the chosen letter.

\paragraph{Biology.}
A union of (i) the biology portion of ScienceQA-style elementary/middle-school
questions, and (ii) the biology subset of SciKnowEval,
which contributes harder graduate-level questions. Both sources are normalized
into a unified schema with a dynamic answer-format specification matching the
actual number of options (2/3/4), so that no item is dropped for being
non-quaternary. The combined dataset therefore contains a mixture of easy and
hard biology MCQs.

\paragraph{LiveNews.}
A recent-news QA dataset built from the human-verified test splits of two
LiveNewsBench~\cite{zhang2026livenewsbench} releases (September 2025 and
January 2026). Each example consists of a free-form factual question about
events that occurred \emph{after} every model's training cutoff, paired
with a short human-written answer. To decouple ``learning the fact'' from
``memorizing the surface phrasing,'' we use \textsc{Gemini3-Flash} to
paraphrase both the question and the answer of every item, producing a
rewritten variant for each original $(q, a)$. \emph{Training is performed on
the rewritten $(q', a')$ pairs, while evaluation is performed on the original
human-written $(q, a)$.} A method that has truly absorbed the underlying news
event must therefore answer a phrasing it has never been trained on, ruling
out trivial token-level memorization.
LiveNews thus measures whether a method can inject genuinely new world
knowledge that cannot have leaked from previous training.

\subsection{Construction of Physics \& Material and Biology}
\label{Construction of Physics}
Both training datasets are reorganized into the common JSONL schema
\{\texttt{prompt}, \texttt{golden\_response}\}.
Every \texttt{prompt} is rendered as a self-contained exam item that bundles
(i) a format specification, (ii) the question stem, and (iii) the option list,
with the format string parameterized by the number of choices $k\in\{2,3,4\}$:
the model is asked to wrap reasoning in
\texttt{<reasoning>...</reasoning>} and to emit a single choice letter inside
\texttt{<answer>...</answer>}.
The corresponding \texttt{golden\_response} follows the same wrapping.

The source of the reference response differs by dataset. ScienceQA-Biology
already provides explanations, so we use them directly and convert the answer index to a letter. SciKnowEval provides only the final answer, so we synthesize reference responses with \textsc{GPT-OSS-20B}: for each item, we sample $N=8$ candidates with $T=0.7$, top-$p=0.9$, and
\texttt{max\_tokens}=1024, discard candidates whose extracted answer letter does not match the gold label, and select the median-length remaining response. Items with no correct candidate are removed. We then form Biology by concatenating ScienceQA-Biology and SciKnowEval-Biology, and
Physics \& Material by concatenating the physics and materials subsets of SciKnowEval.
\section{Experiment Details}
\label{app:exp_details}
\subsection{Baseline Reproduction}

% To make the comparison between SFT, SDFT, and DistillGold
% controlled, we align two implementation choices across all three
% methods: which token positions contribute to the training loss, and
% the direction of the KL divergence used by the distillation methods.

\paragraph{SFT.}
For the SFT baseline we use the \texttt{SFTTrainer} from
TRL~\cite{vonwerra2020trl} with the cross-entropy loss masked to the gold-response
tokens only; the prompt portion of each input is excluded from the
loss support. This matches the loss support used by our GCD and
SDFT.

\paragraph{SDFT.}
For SDFT we run the authors' publicly released
implementation\footnote{\url{https://github.com/idanshen/Self-Distillation}}.
Following a known issue in that repository\footnote{\url{https://github.com/idanshen/Self-Distillation/issues/5}},
we modify the distillation term so that the KL is computed in the
forward direction
$D_{\mathrm{KL}}\!\bigl(\pi_T \,\|\, \pi_\theta\bigr)$
rather than the reverse direction shipped in the original paper.

\subsection{Hyperparameters and Shared Training Pipeline}
All three methods (SFT, SDFT, and GCD) share the same training pipeline for
every (model, task) combination, so that only the loss differs. We use AdamW
with $(\beta_1, \beta_2) = (0.9, 0.999)$, $\varepsilon = 10^{-8}$, and weight
decay $0.01$, clip the global $\ell_2$ gradient norm at $1.0$, and use a linear
warmup followed by a cosine decay of the learning rate to $0$. Each run trains
for $4$ epochs in \texttt{bf16} mixed precision on a single A100 80GB GPU, with a
per-device batch size of $4$ and $8$ gradient accumulation steps (an effective
batch size of $32$) and a maximum sequence length of $2{,}048$ tokens. We perform
full fine-tuning on \mbox{Qwen2.5-3B-Instruct} and Qwen3-4B-Base. On
Llama-3.1-8B-Instruct and Qwen3-8B-Base we use LoRA~\cite{hu2021lora} with rank $r{=}64$ and
scaling factor $128$ on the attention and MLP projections (\texttt{q\_proj},
\texttt{k\_proj}, \texttt{v\_proj}, \texttt{o\_proj}, \texttt{gate\_proj},
\texttt{up\_proj}, \texttt{down\_proj}), keeping all other layers frozen; the
same LoRA configuration is used for all three methods on these models.

\subsection{General Benchmark Evaluation}
\label{app:lmeval}
We evaluate previous capability retention with
\texttt{lm-evaluation-harness} (v0.4.x) using identical settings for the base
model and all fine-tuned checkpoints. All five benchmarks are evaluated
zero-shot. We report accuracy for MMLU and Winogrande, length-normalized
accuracy (\texttt{acc\_norm}) for HellaSwag, MC2 for TruthfulQA, and, for
IFEval, the mean of prompt-level and instruction-level strict accuracy. The
multiple-choice benchmarks are scored by log-likelihood, while IFEval responses
are generated with greedy decoding (temperature $0$) and a maximum of $1{,}280$ new tokens. All
other options follow the harness defaults.

\section{Learning-Rate Sensitivity of SDFT}
\label{app:sdft_lr_ablation}

A natural concern when comparing against SDFT~\cite{shenfeld2026self} is that its
weaker performance might simply reflect a poorly tuned learning rate
rather than a limitation of the method itself. To rule this out, we
conduct a learning-rate sweep for SDFT and contrast every point in the
sweep with a single GCD run that uses the same model, dataset,
optimizer, batch size, schedule, and number of epochs.

For each (model, dataset) combination we train SDFT at five to six
learning rates spanning roughly an order of magnitude---%
$2{\times}10^{-5}$, $5{\times}10^{-5}$, $8{\times}10^{-5}$,
$1{\times}10^{-4}$, $2{\times}10^{-4}$, and $3{\times}10^{-4}$%
\footnote{Llama-3.1-8B-Instruct uses LoRA fine-tuning, so its effective range is shifted
upward; the lowest two rates are skipped for Llama and the highest rate
is skipped for Qwen2.5-3B-Instruct and Qwen3-4B-Base (full fine-tuning). The sweep ranges therefore
match each model's standard tuning regime rather than being held
artificially fixed.}. All other hyperparameters are inherited from
the corresponding configuration in
Appendix~\ref{app:exp_details}; only the learning rate is varied.
For GCD, we plot a single curve trained with the
default reported in Sec.~\ref{sec:sec_4}.

\begin{figure*}[t]
    \centering
    \includegraphics[width=\textwidth]{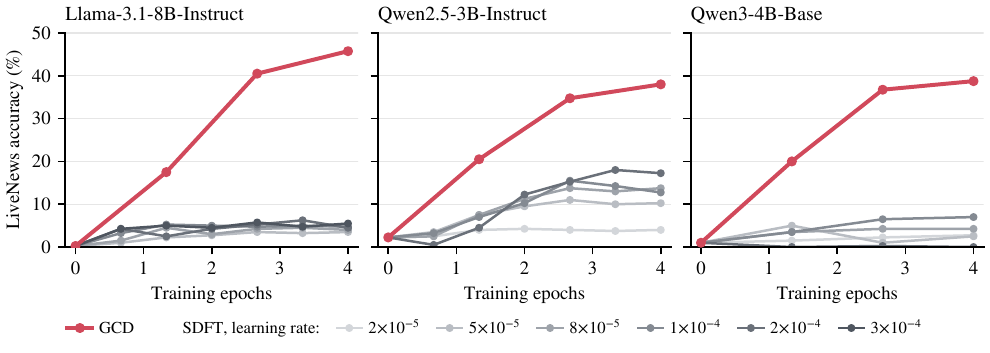}
    \caption{%
        Learning-rate sensitivity of SDFT versus our GCD on LiveNews.
        Each panel shows accuracy across four training epochs for one model. Gray
        curves are SDFT runs, darker for higher learning rates; the red
        curve is GCD at its single reported learning rate. SDFT's best learning rate is model-specific and improves the
        method only marginally, while GCD dominates the entire
        SDFT envelope on every panel.%
    }
    \label{fig:sdft_lr_ablation}
\end{figure*}

\section{EMA Update Rate Ablation}
\label{app:ema_ablation}

GCD maintains the teacher as an exponential moving average of the student,
$\bar{\theta} \leftarrow (1-\alpha)\bar{\theta} + \alpha\theta$, with
$\alpha = 0.01$ in all main experiments. To measure how sensitive GCD is to this
choice, we vary $\alpha \in \{0, 0.01, 0.05, 0.2, 1.0\}$ on Qwen2.5-3B-Instruct
for both LiveNews and Tooluse, keeping every other hyperparameter fixed at the
setting used in the main experiments. Here $\alpha = 0$ freezes the teacher at the
initial model, and $\alpha = 1.0$ copies the student into the teacher after every
update. We report the accuracy of the final checkpoint.

\begin{table}[h]
\centering
\small
\setlength{\tabcolsep}{8pt}
\renewcommand{\arraystretch}{1.15}
\begin{tabular}{lcc}
\toprule
\textbf{EMA rate $\alpha$} & \textbf{LiveNews} & \textbf{Tooluse} \\
\midrule
$0$ (frozen teacher)            & 34.25              & 64.95 \\
$0.01$ (default)                & 38.00              & 66.30 \\
$0.05$                          & 35.75              & 62.89 \\
$0.2$                           & 36.25              & 16.49 \\
$1.0$ (teacher $=$ student)     & 20.50              & \phantom{0}0.00 \\
\bottomrule
\end{tabular}
\caption{\label{tab:ema_ablation}
  Effect of the EMA update rate $\alpha$ on GCD (Qwen2.5-3B-Instruct,
  new-task accuracy in \%, final checkpoint).}
\end{table}

Table~\ref{tab:ema_ablation} shows that GCD is stable for small update rates:
$\alpha \in \{0, 0.01, 0.05\}$ reach similar accuracy on both tasks, so GCD does not
depend on a finely tuned $\alpha$. Large rates cause training to collapse: with
$\alpha = 1.0$ the teacher becomes identical to the student at every step and
accuracy drops sharply on both tasks, and at $\alpha = 0.2$ Tooluse first rises and
then collapses late in training while LiveNews is unaffected. The teacher
therefore needs to lag behind the student, but any small update rate works.

\section{KL Direction Ablation}
\label{app:kl_direction_ablation}

GCD minimizes the forward KL divergence from the teacher to the student,
$D_{\mathrm{KL}}(\pi_{\bar{\theta}} \,\|\, \pi_{\theta})$ (Sec.~\ref{sec:method}). To test this choice, we replace it with the reverse KL
divergence, $D_{\mathrm{KL}}(\pi_{\theta} \,\|\, \pi_{\bar{\theta}})$, on
Qwen2.5-3B-Instruct for both LiveNews and Tooluse, keeping every other
hyperparameter fixed at the setting used in the main experiments. We report the
accuracy of the final checkpoint.

\begin{table}[h]
\centering
\small
\setlength{\tabcolsep}{8pt}
\renewcommand{\arraystretch}{1.15}
\begin{tabular}{lcc}
\toprule
\textbf{KL direction} & \textbf{LiveNews} & \textbf{Tooluse} \\
\midrule
Forward (default) & 38.00 & 66.30 \\
Reverse           & 11.00 & 55.67 \\
\bottomrule
\end{tabular}
\caption{\label{tab:kl_direction_ablation}
  Effect of the KL direction on GCD (Qwen2.5-3B-Instruct, new-task accuracy in
  \%, final checkpoint).}
\end{table}

Table~\ref{tab:kl_direction_ablation} shows that forward KL outperforms reverse KL
on both tasks. The gap is largest on LiveNews, where the target knowledge is
absent before fine-tuning: with reverse KL the model barely learns the new facts,
reaching only 11.00\% compared with 38.00\% for forward KL. On Tooluse, where the
base model already has partial ability (34.02\%), reverse KL still learns the task
but trails forward KL by 10.63 points (55.67\% vs.\ 66.30\%).

\section{Multi-Seed Results}
\label{app:multi_seed}

\ifdefined\arxivversion\begin{table*}[H]\else\begin{table*}[t]\fi
\centering
\setlength{\tabcolsep}{8pt}
\renewcommand{\arraystretch}{1.15}
\begin{tabular}{lccc}
\toprule
\textbf{Method} & \textbf{Qwen2.5-3B-Instruct} & \textbf{Qwen3-4B-Base} & \textbf{Llama-3.1-8B-Instruct} \\
\midrule
SFT  & 61.1 $\pm$ 0.7 & 63.2 $\pm$ 0.6 & 66.0 $\pm$ 0.4 \\
SDFT & 56.4 $\pm$ 0.6 & 61.2 $\pm$ 0.8 & 35.6 $\pm$ 0.7 \\
\textbf{GCD} & \textbf{66.7 $\pm$ 0.6} & \textbf{69.6 $\pm$ 0.5} & \textbf{66.9 $\pm$ 0.4} \\
\bottomrule
\end{tabular}
\caption{\label{tab:multi_seed}
  Tooluse accuracy (\%) over five random seeds, reported as mean $\pm$ standard
  deviation.}
\end{table*}

To check that the gains of GCD do not come from a single favorable run, we repeat
the Tooluse experiments with five random seeds on three of the four models
(Qwen2.5-3B-Instruct, Qwen3-4B-Base, and Llama-3.1-8B-Instruct), keeping every
other hyperparameter fixed at the setting used in the main experiments. Given the
training cost of multi-seed evaluation, we use Tooluse as a representative task.
Table~\ref{tab:multi_seed} reports the mean and standard deviation of the
new-task accuracy over the five seeds.

GCD achieves the highest mean accuracy on all three models: 66.7\% on
Qwen2.5-3B-Instruct, 69.6\% on Qwen3-4B-Base, and 66.9\% on Llama-3.1-8B-Instruct.
The standard deviations of all methods are below one point, much smaller than the
gaps between GCD and SDFT (10.3, 8.4, and 31.3 points) and between GCD and SFT on
the two Qwen models (5.6 and 6.4 points). The improvements of GCD are therefore
consistent across random seeds.

\end{document}